\documentclass[runningheads]{llncs}
\pdfoutput=1
\usepackage[utf8]{inputenc}
\usepackage{color}
\usepackage{array}
\usepackage{booktabs}
\usepackage{multirow}
\usepackage{amsmath}
\usepackage{amssymb}
\usepackage{graphicx}
\usepackage{setspace}
\usepackage[unicode=true,pdfusetitle,
 bookmarks=true,bookmarksnumbered=false,bookmarksopen=false,
 breaklinks=false,pdfborder={0 0 1},backref=false,colorlinks=false]
 {hyperref}

\makeatletter

\providecommand{\tabularnewline}{\\}
\newcommand{\lyxdot}{.}

\newcommand{\etal}{\textit{et al.}}
\usepackage{graphicx}
\usepackage{amsmath,amssymb} 
\usepackage{color}

\usepackage[english]{babel}

\@ifundefined{showcaptionsetup}{}{%
 \PassOptionsToPackage{caption=false}{subfig}}
\usepackage{subfig}
\makeatother

\begin{document}
\pagestyle{headings} \mainmatter



\title{Visual Concept Recognition and Localization via Iterative Introspection}

\author{Amir Rosenfeld $\qquad$Shimon Ullman}

\institute{Department of Computer Science and Applied Mathematics\\
Weizmann Institute of Science\\
Rehovot, Israel}

\maketitle

\begin{abstract}
Convolutional neural networks have been shown to develop internal
representations, which correspond closely to semantically meaningful
objects and parts, although trained solely on class labels. Class
Activation Mapping (CAM) is a recent method that makes it possible
to easily highlight the image regions contributing to a network's
classification decision. We build upon these two developments to enable
a network to re-examine informative image regions, which we term \emph{introspection}.
We propose a weakly-supervised iterative scheme, which shifts its
center of attention to increasingly discriminative regions as it progresses,
by alternating stages of classification and introspection. We evaluate
our method and show its effectiveness over a range of several datasets,
where we obtain competitive or state-of-the-art results: on Stanford-40
Actions, we set a new state-of the art of 81.74\%. On FGVC-Aircraft
and the Stanford Dogs dataset, we show consistent improvements over
baselines, some of which include significantly more supervision.

\end{abstract}

\section{Introduction}

\begin{figure}
\begin{centering}
\subfloat[\textcolor{red}{brushing teeth}]{\includegraphics[width=0.25\linewidth,height=0.25\linewidth]{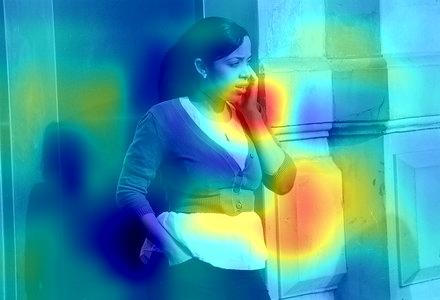}

}\subfloat[\textcolor{red}{looking through a telescope}]{\includegraphics[width=0.25\linewidth,height=0.25\linewidth]{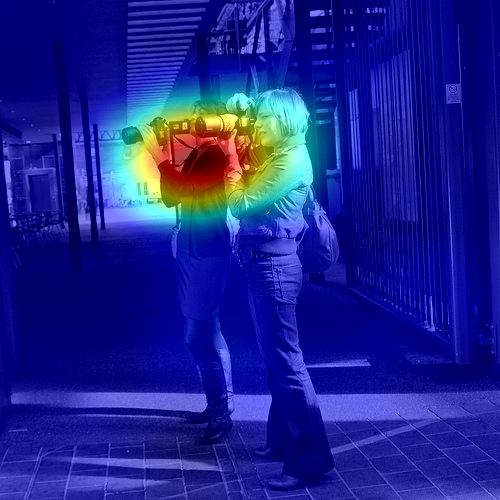}}\subfloat[\textcolor{red}{fishing}]{\includegraphics[width=0.25\linewidth,height=0.25\linewidth]{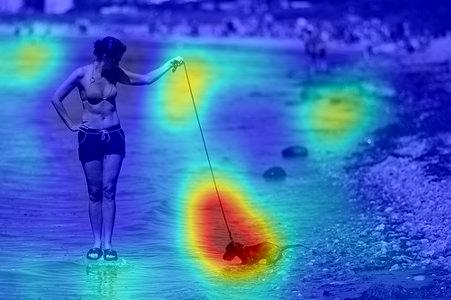}}
\par\end{centering}

\centering{}\subfloat[\textcolor{green}{phoning}]{\includegraphics[width=0.25\linewidth,height=0.25\linewidth]{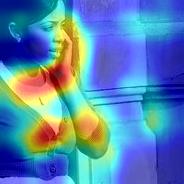}

}\subfloat[\textcolor{green}{taking photo}]{\includegraphics[width=0.25\linewidth,height=0.25\linewidth]{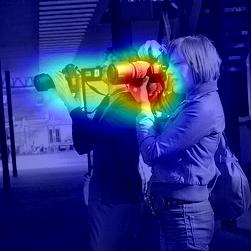}}\subfloat[\textcolor{green}{walking a dog}]{\includegraphics[width=0.25\linewidth,height=0.25\linewidth]{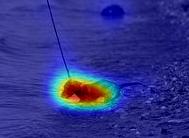}}\protect\caption{\label{fig:(Top)-Class-Activation}(\emph{Top}) Class Activation Maps
\cite{zhou2015learning} show the source of a network's classification.
The network tends to focus on relevant image regions even if its final
prediction is wrong. An SVM trained on features extracted from VGG-GAP
\cite{zhou2015learning} misclassified all of these images, while
highlighting the discriminative regions. (\emph{a,b,c}) the predicted
classes appears in red. (\emph{Bottom, zoomed in version of top})
The proposed method effectively removes many such errors by focusing
attention on the highlighted regions. (\emph{d,e,f}) the corrected
prediction following the introspection stage appears in green.}
\end{figure}

With the advent of deep convolutional neural networks as the leading
method in computer vision, several attempts have been made to understand
their inner workings. Examples of pioneering work in this direction
include \cite{zeiler2014visualizing,mahendran2015understanding};
providing glimpses into the representations learned by intermediate
levels in the network. Specifically, the recent work of Zhou \etal\
\cite{zhou2015learning} provides an elegant mechanism to highlight
the discriminative image regions that served the CNN for a given task.
This can be seen as a form of \textit{\emph{introspection}}, highlighting
the source of the network's conclusions. A useful trait we have observed
in experiments is that even if the final classification is incorrect,
the highlighted image regions are still be informative with respect
to the correct target class. This is probably due to the similar appearance
of confused classes. see Figures \ref{fig:(Top)-Class-Activation}
and \ref{fig:Class vs location} for some examples. Motivated by this
observation, we propose an iterative mechanism of internal supervision,
termed introspection, which revisits discriminative regions to refine
the classification. As the process is repeated, each stage further
highlights discriminative sub-regions. Each stage uses its own classifier,
as we found this to be beneficial when compared to using the same
classifier for all sub-windows. 

We describe strategies for how to leverage the introspection scheme,
and demonstrate how these consistently improve results on several
benchmark datasets, while progressively refining the localization
of discriminative regions. As shown, our method is particularly beneficial
for fine-grained tasks such as species \cite{wah2011caltech,KhoslaYaoJayadevaprakashFeiFei_FGVC2011}
or model \cite{maji2013fine} identification and to challenging cases
in e.g., action recognition \cite{yao2011human}, which requires attention
to small and localized details. 

In the following we will first review some related work. In Section
\ref{sec:blind} we describe our method in detail. Section \ref{sec:Experiments}
contains experiments and analysis to evaluate the proposed method,
followed by concluding remarks in \ref{sec:Conclusions}.

\section{Related Work}

Supervised methods consistently outperform unsupervised or semi-supervised
methods, as they allow for the incorporation of prior knowledge into
the learning process. There is a trade-off between more accurate classification
results and structured output on the one and, the cost of labor-intensive
manual annotations, on the other. Some examples are \cite{zhang2014part,zhang2015fine},
where bounding boxes and part annotations are given at train time.
Aside from the resources required for large-scale annotations, such
methods elude the question of learning from weakly supervised data
(and mostly unsupervised data), as is known to happen in human infants,
who can learn from limited examples \cite{lake2015human}. Following
are a few lines of work related to the proposed method.

\subsection{Neural Net Visualization and Inversion}

Several methods have been proposed to visualize the output of a neural
net or explore its internal activations. Zeiler \etal\ \cite{zeiler2014visualizing}
found patterns that activate hidden units via deconvolutional neural
networks. They also explore the localization ability of a CNN by observing
the change in classification as different image regions are masked
out. \cite{mahendran2015understanding} Solves an optimization problem,
aiming to generate an image whose features are similar to a target
image, regularized by a natural image prior. Zhou \etal\ \cite{zhou2014object}
aims to explicitly find what image patches activate hidden network
units, finding that indeed many of them correspond to semantic concepts
and object parts. These visualizations suggest that, despite training
solely with image labels, there is much to exploit within the internal
representations learned by the network and that the emergent representations
can be used for weakly supervised localization and other tasks of
fine-grained nature.

\subsection{Semi-Supervised class Localization}

Some recent works attempt to obtain object localization through weak
labels, i.e., the net is trained on image-level class labels, but
it also learns localization. \cite{bergamo2014self} Localizes image
regions pertaining to the target class by masking out sub-images and
inspecting change in activations of the network. Oquab \etal\ \cite{oquab2015object}
use global max-pooling to obtain points on the target objects. Recently,
Zhou \etal\ \cite{zhou2015learning} used global average pooling
(GAP) to generate a Class-Activation Mapping (CAM), visualizing discriminative
image regions and enabling the localization of detected concepts.
Our introspection mechanism utilizes their CAMs to iteratively identify
discriminative regions and uses them to improve classification without
additional supervision.

\subsection{Attention Based Mechanisms}

Recently, some attention based mechanisms have been proposed, which
allow focusing on relevant image regions, either for the task of better
classification \cite{mnih2014recurrent} or efficient object localization
\cite{caicedo2015active}. Such methods benefit from the recent fusion
between the fields of deep learning and reinforcement learning \cite{mnih2013playing}.
Another method of interest is the spatial-transformer networks in
\cite{jaderberg2015spatial}: they designed a network that learns
and applies spatial warping to the feature maps, effectively aligning
inputs, which results in increased robustness to geometric transformations.
This enables fine-grained categorization on the CUB-200-2011 birds
\cite{wah2011caltech} dataset by transforming the image so that only
discriminative parts are considered (bird's head, body). Additional
works appear in \cite{xiao2015application}, who discovers discriminative
patches and groups them to generate part detectors, whose detections
are combined with the discovered patches for a final classification.
In \cite{lin2015bilinear}, the outputs of two networks are combined
via an outer-product, creating a strong feature representation. \cite{krause2015fine}
discovers and uses parts by using co-segmentation on ground-truth
bounding boxes followed by alignment. 

\begin{figure}
\centering{}\includegraphics[width=0.82\textwidth]{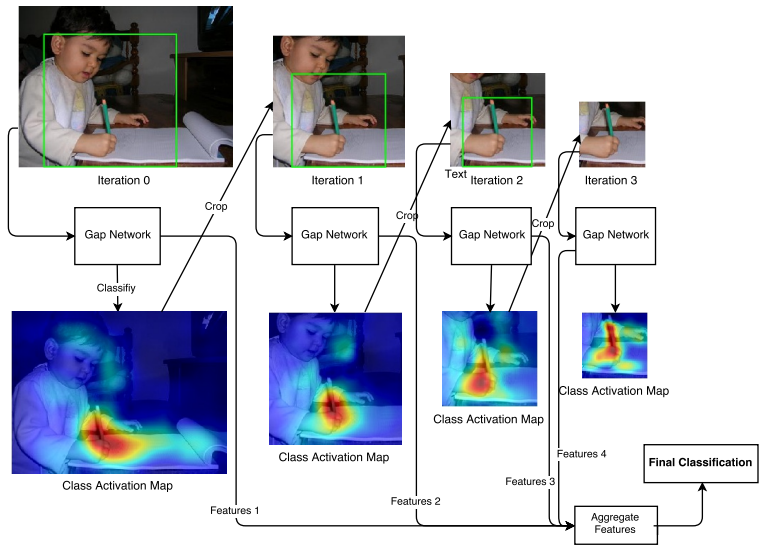}\protect\caption{\label{fig:Overview-of-proposed}Overview of proposed method. At each
iteration, an image window (\emph{top}) is classified using features
from a GAP-network. Top scoring classes are used to generate Class
Activation Maps (\emph{bottom}), which are then used to select sub-windows
(green rectangles). The process is repeated for a few iterations and
features from all visited image windows are aggregated and used in
a final classifier. The correct class is ``writing on a book''.
Attention shifts gradually and closes in on the discriminative image
region, i.e., the boy's hand holding a pencil.}
\end{figure}

\section{Approach\label{sec:blind}}

Our approach is composed of alternating between two main steps: classification
and introspection. In the \emph{classification} step, we apply a trained
network to an image region (possibly the entire image). In the \emph{introspection}
step, we use the output of a hidden layer in the network, whose values
were set during the classification step. This highlights image regions
which are fed to the next iteration's classification step. This process
is iterated a few times (typically 4, see Section \ref{sec:Experiments},
Fig. \ref{fig:Early-vs.-Late}), and finally the results of all stages
are combined. Training proceeds by learning a specialized classifier
for each iteration, as different iterations capture different contexts
and levels of detail (but without additional supervision). 

	Both classification/introspection steps utilize the recent Class-Activation
Mapping method of \cite{zhou2015learning}. We briefly review the
CAM method and then describe how we build upon it. In \cite{zhou2015learning},
a deep neural network is modified so that post-classification, it
is possible to visualize the varying contribution of image regions,
via a so-called Class Activation Mapping (CAM). A global average pooling
was used as a penultimate feature representation. This results in
a feature vector which is a spatial averaging of each of the feature
maps of the last convolutional layer. Using the notation in \cite{zhou2015learning}:
let $f_{k}(x,y)$ be the $k$'th output of the last convolutional
layer at grid location $(x,y)$. The results of the global-average
pooling results in a vector $F=(F_{1},F_{2},\dots,F_{k})$, defined
as:

\begin{equation}
F_{k}=\sum_{x,y}f_{k}(x,y)
\end{equation}

This is followed by a fully connected layer with $C$ outputs (assuming
$C$ target classes). Hence the score for class $c$ before the soft-max
will be:

\begin{singlespace}
\begin{align}
S_{c} & =\sum_{k}w_{k}^{c}F_{k}\\
 & =\sum_{k}w_{k}^{c}\sum_{x,y}f_{k}(x,y)\\
 & =\sum_{x,y}\sum_{k}w_{k}^{c}f_{k}(x,y)
\end{align}

\end{singlespace}

Now, define 

\begin{equation}
M_{c}(x,y)=\sum_{k}\omega_{k}^{c}f_{k}(x,y)\label{eq:contribution}
\end{equation}

where $\omega_{k}^{c}$ are class-specific weights. Hence we can express
$S_{c}$ as a summation of terms over $(x,y)$: 

\begin{equation}
S_{c}=\sum_{x,y}M_{c}(x,y)\label{eq:spatial weighting}
\end{equation}

And the class probability scores are computed via soft-max, e.g, $P_{c}=\frac{e^{S_{c}}}{\sum_{t}e^{S_{t}}}$.
Eq. \ref{eq:contribution} allows us to measure the contribution of
each grid cell $M_{c}(x,y)$ \emph{for each specific class }$c$.
Indeed, \cite{zhou2015learning} has shown this method to highlight
informative image regions (with respect to the task at hand), while
being on par with the classification performance obtained by the unmodified
network (GoogLeNet \cite{szegedy2015going} in their case). See Fig.
\ref{fig:(Top)-Class-Activation} for some CAMs. Interestingly, we
can use the CAM method to highlight informative image regions for
classes other than the correct class, providing intuition on the features
it has learned to recognize. This is discussed in more detail in Section
\ref{sub:Correlation-of-Class} and demonstrated in Fig. \ref{fig:Class vs location}.
We name a network whose final convolutional layer is followed by a
GAP layer as a GAP-network, and the output of the GAP layer as the
GAP features. We next describe how this is used in our proposed method.

\subsection{\label{sub:Iterative-Classification-Introsp}Iterative Classification-Introspection}

The proposed method alternates classification and introspection. Here
we provide the outline of the method, with specific details such as
values of parameters discussed in Section \ref{sub:Various-Parameters-=000026}. 

For a given image $I$ and window $w$ (initially the entire image),
a learned classifier is applied to the GAP features extracted from
the window $I_{w}$, resulting in $C$ classification scores $S_{c}$,
$c\in[1\dots C]$ and corresponding CAMs $M_{c}^{w}(x,y)$. The introspection
phase employs a strategy to select a sub-window for the next step
by applying a beam-search to a set of putative sub-windows. The sequence
of windows visited by the method is a route on an \emph{exploration-tree},
from the root to one of the leaves. Each node represents an image
window and the root is the entire image. We next explain how the sub-windows
are created, and how the search is applied. 

We order the current classification scores $S_{c}$ by descending
order and retain the top $k$ scoring classes. Let $\hat{c}$ be one
of these classes and $M_{\hat{c}}^{w}(x,y)$ the corresponding CAM.
We extract a square sub-window $w'$ centered on the maximal value
of $M_{\hat{c}}^{w}(x,y)$. Each such $w'$ ($k$ in total) is added
as a child of the current node, which is represented by $w$. In this
way, each iteration of the method expands a selected node in the exploration-tree,
corresponding to an image window, until a maximum depth is reached.
The tree depth is the number of iterations. We define iteration $0$
as the iteration acting on the root. The size of a sub-window $w'$
is of a constant fraction of the size of its parent $w$. We next
describe how the exploration-tree is used for classification.

\subsection{Feature Aggregation}

Let $k$ be the number of windows generated at iteration $t>0$. We
denote the set of windows by: 
\begin{equation}
W_{t}=(w_{i}^{t})_{i=1}^{k}
\end{equation}
And the entire set of windows as:

\begin{equation}
\mathcal{R}=(W_{t})_{t=0}^{T}
\end{equation}

where $W_{0}$ is the entire image. For each window $w_{i}^{t}$ we
extract features $f_{w}^{t}\in\mathbb{R}^{K}$, e.g., $K=1024$, the
dimension of the GAP features, as well as classification scores $S_{w_{i}^{t}}\in\mathbb{R}^{C}$.
The set of windows $\mathcal{R}$ for an image $I$ is arranged\emph{
}as nodes in the exploration-tree. The final prediction is a result
of aggregating evidence from selected sub-windows along some path
from the root to a tree-leaf. We evaluate variants of both early fusion
(combining features from different iterations) or later fusion (combining
predictions from different iterations).

\subsection{Training}

Training proceeds in two main stages. The first is to train a sequence
of classifiers that will produce an exploration-tree for each training/testing
sample. The second is training on feature aggregations along different
routes in the exploration-trees, to produce a final model. 

During training, we train a classifier for each iteration (for a predefined
number of iterations, 5 total) of the introspection/classification
sequence. The automatic training of multiple classifiers at different
scales contributes directly to the success of the method, as using
the same classifier for all iterations yielded no improvement over
the baseline results (Section \ref{sub:Setup:}). For the first iteration,
we simply train on entire images with the ground-truth class-labels.
For each iteration $t>1$, we set the training samples to sub-windows
of the original images and the targets to the ground-truth labels.
The sub-windows selected for training are always those corresponding
to the strongest local maximum in $M_{c}(x,y)$, where $M_{\hat{c}}(x,y)$
is the CAM corresponding to the highest scoring class. Each classifier
is an SVM trained on the features the output of the GAP layer of the
network (as was done in \cite{zhou2015learning}). We also checked
the effect of fine-tuning the network and using additional features.
The Results are discussed in the experiments, Section \ref{sec:Experiments}.

\begin{center}
\begin{figure}
\begin{centering}
\begin{minipage}[t]{1\columnwidth}%
\includegraphics[width=1\textwidth]{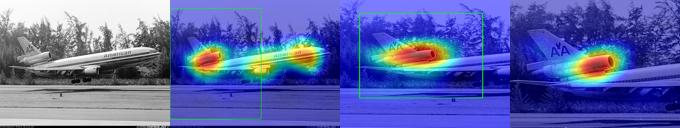}%
\end{minipage}
\par\end{centering}

\smallskip{}

\begin{centering}
\begin{minipage}[t]{1\columnwidth}%
\includegraphics[width=1\textwidth]{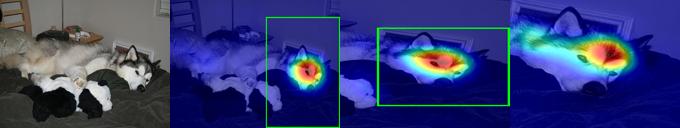}%
\end{minipage}
\par\end{centering}

\smallskip{}

\begin{centering}
\begin{minipage}[t]{1\columnwidth}%
\includegraphics[width=1\textwidth]{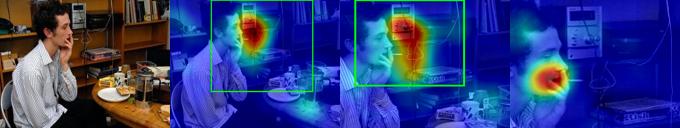}%
\end{minipage}
\par\end{centering}

\protect\caption{\label{fig:Exploration-Routes-on}Exploration routes on images: Each
row shows the original image and 3 iterations of the algorithm, including
the resulting Class-Activation Maps \cite{zhou2015learning} used
to guide the next iteration. The selected sub-window is shown with
a green bounding box. Despite being trained and tested without any
bounding box annotations, the proposed method closes in on the features
relevant to the target class. The first 2 predictions (\emph{columns
2,3}) in each row are mistaken and the last one (\emph{rightmost column})
is correct.}
\end{figure}

\par\end{center}

\subsubsection*{Routes on Exploration Trees}

The image is explored by traversing routes on a tree of nested sub-windows.
The result of training is a set of classifiers, $\mathcal{E}=(E)_{i=1\dots T}$.
We produce an exploration tree by applying at each iteration $j$
the classifier $E_{j}$ on the features of the windows produced by
the previous iteration. The window of iteration $0$ is the entire
image. A route along the tree will consist of a sequence of windows
$w^{1},w^{2},...w^{T}$ where T is the number of iterations. We found
in experiments that more than 5 iterations (including the first) brings
negligible boosts in performance. The image score for a given class
is given by either (1) summing the scores of classifiers along a route
(late fusion), or (2) learning a classifier for the combined features
of all visited windows along the route (early fusion). Features are
combined via averaging rather than concatenation. This reduces the
training time at no significant change to the final performance; such
an effect has also been noted by \cite{simonyan2014very}. See Fig.
\ref{fig:Overview-of-proposed} for an overview of the proposed method.
Fig. \ref{fig:Exploration-Routes-on} shows some examples of how progressively
zooming in on image regions helps correct early classification mistakes. 

\begin{table}
\protect\caption{\label{tab:Classification-Accuracy-of}Classification accuracy of
our method vs. a baseline for several datasets. VGG-GAP{*} is our
improved baseline using the VGG-GAP network \cite{zhou2015learning}.
ours-late-fusion: we aggregate the scores of image windows along the
visited path. ours-early-fusion: aggregate scores of classifiers trained
on feature combination of windows along the visited path. ours-ft:
same as ours-early fusion but fine-tuned. +D: concatenated with fc6
features from VGG-16 at each stage. {*}Stanford Dogs is a subset of
ILSVRC dataset. For this dataset, we compare to work which also used
a network pre-trained on ILSVRC. $^{\dagger}$means fine tuning the
network on all iterations.}
\begin{tabular*}{1\textwidth}{@{\extracolsep{\fill}}l>{\raggedright}p{0.13\textwidth}>{\raggedright}p{0.1\textwidth}>{\raggedright}p{0.2\textwidth}l}
\toprule 
\textbf{Method} & \textbf{40-Actions } & \textbf{Dogs}{*} & \textbf{Birds}  & \textbf{Aircraft}\tabularnewline
\midrule 
GoogLeNet-GAP\cite{zhou2015learning} & $72.03$ & - & $63.00$ & -\tabularnewline
\midrule 
VGG-16-fc6 & $73.83$ & $83.76$ & $63.46$ & $60.07$\tabularnewline
\midrule 
VGG-GAP{*} & $75.31$ & $81.83$ & $65.72$ & $62.95$\tabularnewline
\midrule 
\multirow{1}{*}{ours-late-fusion} & $76.88$ & $82.63$ & $73.52$ & $66.34$\tabularnewline
\midrule 
ours-early-fusion & $77.08$ & $83.55$ & $71.64$ & $68.26$\tabularnewline
\midrule 
ours-ft & $80.37$ & $82.62$ & $78.74$ & $79.15$\tabularnewline
\midrule 
ours-early-fusion+D & $81.04$ & \textbf{$\mathbf{86.25}$} & $78.91$ & \textbf{$77.74$}\tabularnewline
\midrule 
ours-ft+D & \textbf{$\mathbf{81.74}$} & $84.18$ & \multicolumn{1}{l}{$79.55$} & \textbf{$78.04$}\tabularnewline
\midrule 
Previous Work & $72.03$\cite{zhou2015learning},$81$

\cite{gao2015deep} & $79.92$\cite{zhang2015weakly} & $77.9$\cite{xiao2015application}, $81.01$ \cite{simon2015neural},
$82$ \cite{krause2015fine}, \textbf{$\mathbf{84.1}$ }\cite{jaderberg2015spatial} & \textbf{$\mathbf{84.1}$\cite{lin2015bilinear}}\tabularnewline
\bottomrule
\end{tabular*}
\end{table}

\section{Experiments\label{sec:Experiments}}

\subsection{\label{sub:Setup:}Setup: }

In all our experiments, we start with a variant of the VGG-16 network
\cite{simonyan2014very} which was fined tuned on ILSVRC by \cite{zhou2015learning}.
We chose it over GoogLeNet-GAP as it obtained slightly higher classification
results on the ILSVRC validation set. In this network, all layers
after \emph{conv5-3} have been removed, including the subsequent pooling
layer; hence the spatial-resolution of the resultant feature maps/CAM
is $14\times14$ for an input of size $224\times224$ (leaving the
pooling layer would reduce resolution to be $7\times7$). A convolutional
layer of $1024$ filters has been added, followed by a fully-connected
layer to predict classes. This is our basic GAP-network, called VGG-GAP.
Each image window, including the entire original image, is resized
so that its smaller dimension is $224$ pixels, resulting in a feature
map $14\times n\times1024$, for which we compute the average along
the first two dimensions, to get a feature representation. We resize
the images using bilinear interpolation. We train a separate classifier
for each iteration of the classification/introspection process; treating
all visited image windows with the same classifier yielded a negligible
improvement (0.3\% in precision) over the baseline. All classifiers
are trained with a linear SVM \cite{liblinear} on $\ell_{2}$ normalized
feature vectors. If the features are a concatenation of two feature
vectors, those are $\ell_{2}$ normalized before concatenation. Our
experiments were carried out using the MatConvNet framework \cite{vedaldi2015matconvnet},
as well as \cite{PMT,conf/mm/VedaldiF10}. We evaluated our approach
on several datasets, including Stanford-40 Actions \cite{yao2011human},
the Caltech-USCD Birds-200-2011 \cite{wah2011caltech} (a.k.a CUB-200-2011),
the Stanford-Dogs dataset, \cite{KhoslaYaoJayadevaprakashFeiFei_FGVC2011}
and the FGVC-Aircraft dataset \cite{maji2013fine}. See Table \ref{tab:Classification-Accuracy-of}
for a summary of our results compared to recent work. In the following,
we shall first show some analysis on the validity of using the CAMs
to guide the next step. We shall then describe interesting properties
of our method, as well as the effects of different parameterizations
of the method.

\subsection{\label{sub:Correlation-of-Class}Correlation of Class and Localization}

In this section, we show some examples to verify our observation that
CAMs tend to highlight informative image locations w.r.t to the target
class despite the fact that the image may have been misclassified
at the first iteration (i.e., before zooming in on sub-windows).

\begin{figure}
\begin{centering}
\begin{minipage}[t]{0.49\columnwidth}%
\includegraphics[clip,width=1\textwidth]{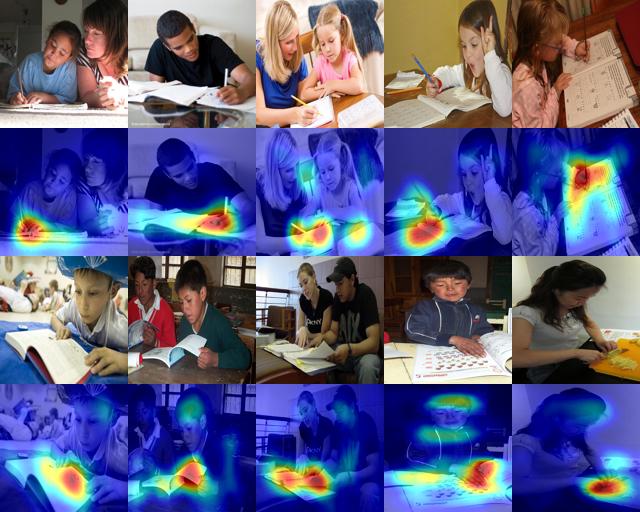}

\begin{center}
(writing on a book)
\par\end{center}%
\end{minipage}\hfill{}%
\begin{minipage}[t]{0.49\columnwidth}%
\includegraphics[clip,width=1\textwidth]{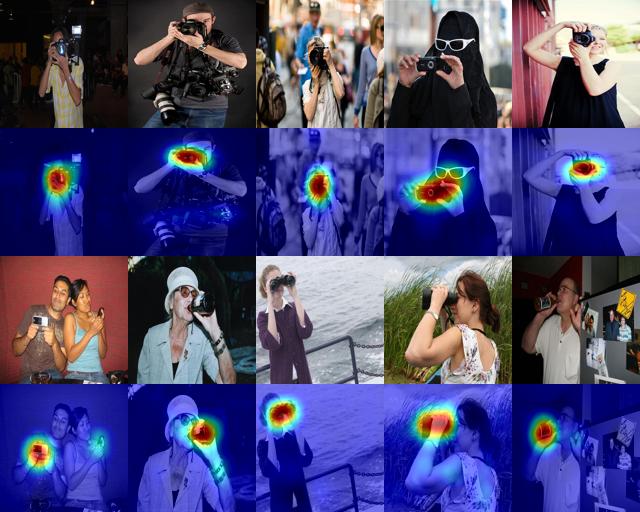}

\begin{center}
(taking photos)
\par\end{center}%
\end{minipage}
\par\end{centering}

\begin{centering}
\begin{minipage}[t]{0.49\columnwidth}%
\includegraphics[clip,width=1\textwidth]{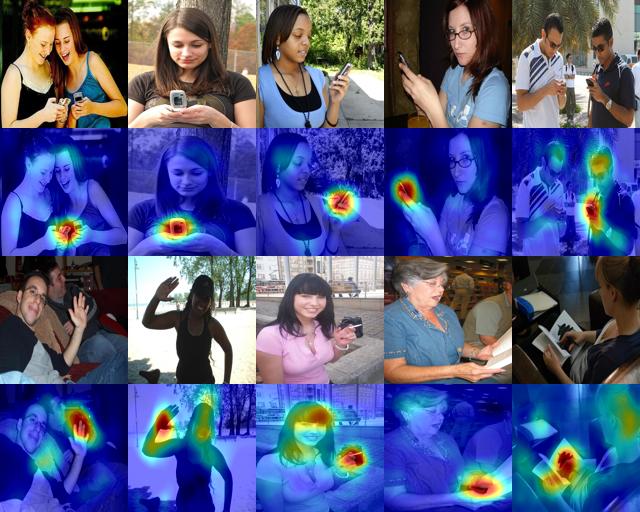}

\begin{center}
(texting message)
\par\end{center}%
\end{minipage}\hfill{}%
\begin{minipage}[t]{0.49\columnwidth}%
\includegraphics[clip,width=1\textwidth]{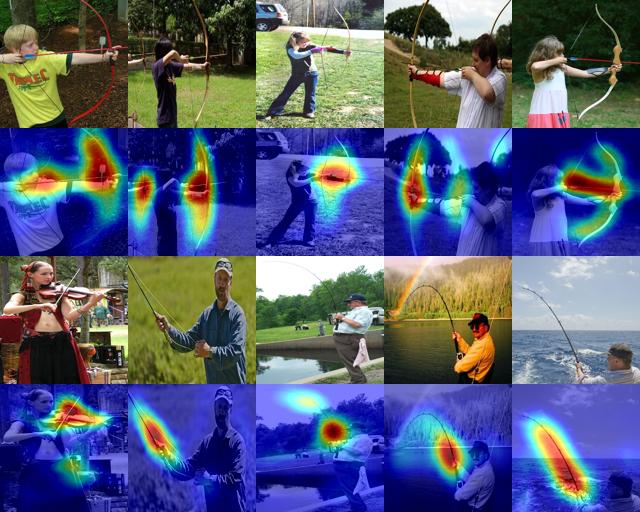}

\begin{center}
(shooting an arrow)
\par\end{center}%
\end{minipage}
\par\end{centering}

\protect\caption{\label{fig:Class vs location}Top ranking images (both true positives
and false positives) for various action categories along with Class-Activation
Maps. Misclassified images still carry information on where additional
attention will disambiguate the classification. Each block of images
shows, from first to fourth row: high-ranking true-positives and their
respective CAMs , high ranking false-positive and their respective
CAMs. The target class appears below each block. We recommend viewing
this figure online to zoom in on the details.}
\end{figure}

To do so, we have applied to the test-set of the Stanford-40 actions
dataset a classifier learned on the GAP features of VGG-GAP. For each
category in turn, we ranked all images in the test set according to
the classifier's score for that category. We then picked the top $5$
true positive images and top $5$ false positive images. See Fig.
\ref{fig:Class vs location} for some representative images. We can
see that the CAMs for a target class tend to be consistent in both
positive images and high-ranking non-class images. Intuitively, this
is because the classifier gives more weight to patterns which are
similar in appearance. For the ``writing on a book'' category (top-left
of Fig. \ref{fig:Class vs location} ) we can see how in positive
images the books and especially the hands are highlighted, as they
are for non-class images, such as ``reading'', or ``cutting vegetables''.
For ``texting message'' (top-right) the hand region is highlighted
in all images, regardless of class. For ``shooting an arrow'' class
(bottom-right), elongated structures such as fishing rods are highlighted.
A nice confusion appears between an archer's bow and a violinist's
bow (bottom-right block, last row, first image), which are also referred
to by the same word in some human languages.

To check our claim quantitatively, we computed the extent of two square
sub-windows for each image in the test-set: one using the CAM of the
true class and one using the CAM of the highest scoring non-true class.
For each pair we computed the overlap (intersection over union) score.
The mean score of all images was $0.638$; This is complementary evidence
to \cite{zhou2015learning}, who shows that the CAMs have good localization
capabilities for the correct class.

\subsection{Early and Late Fusion}

\begin{figure}
\centering{}\includegraphics[height=0.25\paperheight]{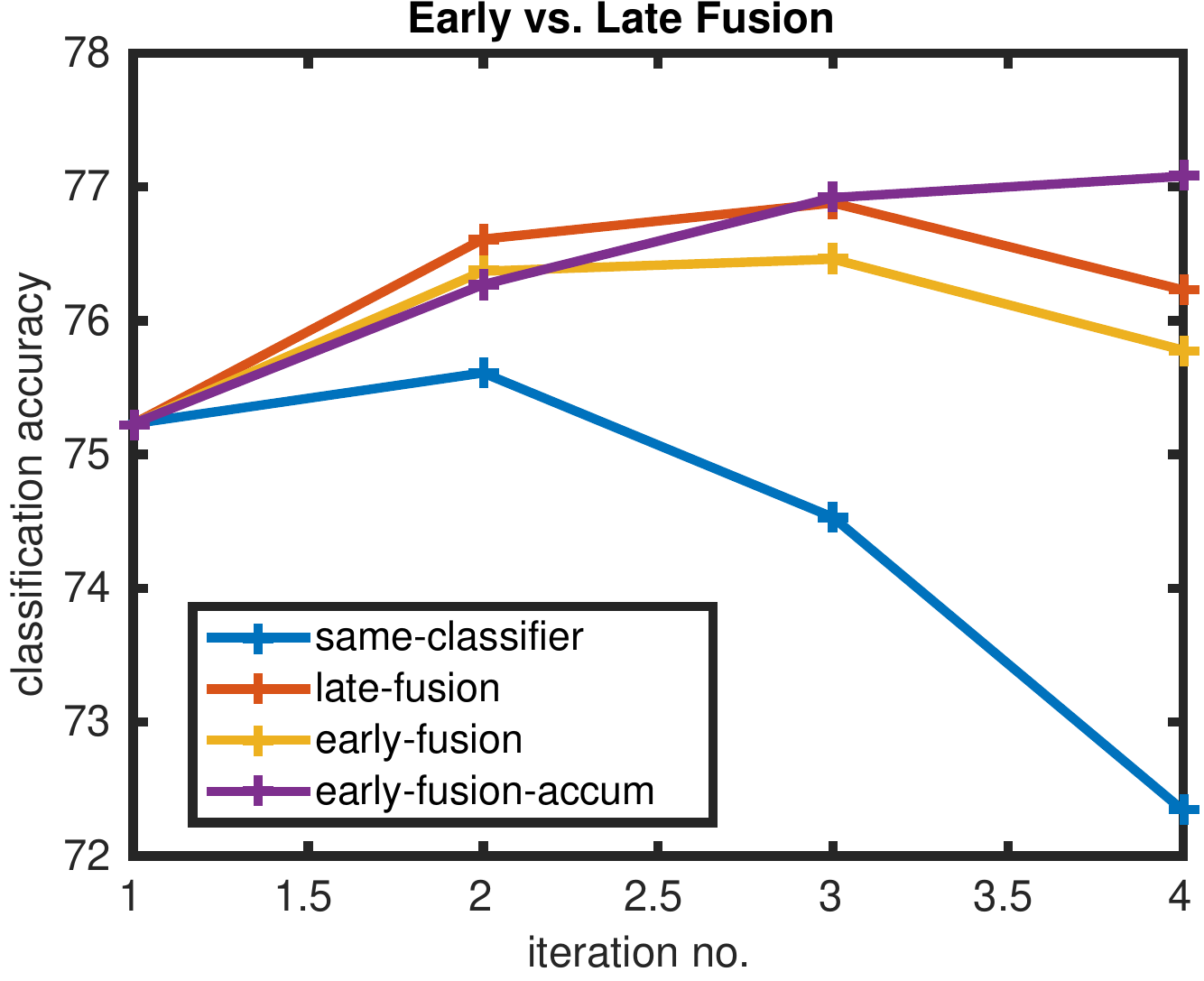}\protect\caption{\label{fig:Early-vs.-Late}Effect of learning at different iterations:
using the same classifier trained on entire images for all iterations
tends to cause overall precision to drop (\emph{same-classifier}).
Accumulating the scores of per-iteration learned classifiers along
the explored path improves this (\emph{late-fusion}). Using all features
along the observed exploration path improves classification as the
path length increases (\emph{early-fusion}) and summing all scores
along the early-fusion path brings the best performance (\emph{early-fusion-accum}).
Performance is shown on the Stanford-40 \cite{yao2011human} dataset}
\end{figure}

In all our experiments, we found that using the features extracted
from a window at some iteration can bring worse results on its own
compared to those extracted from earlier iterations (which include
this window). However, the performance tends to improve as we combine
results from several iterations, in a late-fusion manner. Training
on the combined (averaged) features of windows from multiple iterations
further improves the results (early-fusion). Summing the scores of
early-fused features for different route lengths further improves
accuracy: if $S_{i}$ is the score of the classifier trained on a
route of length $i$. Then creating a final score from $S_{1}+S_{t}+\dots S_{T}$
tends to improve as $T$ grows, typically stabilizing at $T=5$. See
Fig. \ref{fig:Early-vs.-Late} for an illustration of this effect.
Importantly, we tried using the classifier from the first iteration
(i.e., trained on entire images) for all iterations. This performed
worse than learning a classifier per-iteration, especially in later
iterations.

\subsection{Fine-Grained vs General Categories}

\begin{figure}
\centering{}\includegraphics[width=0.7\textwidth]{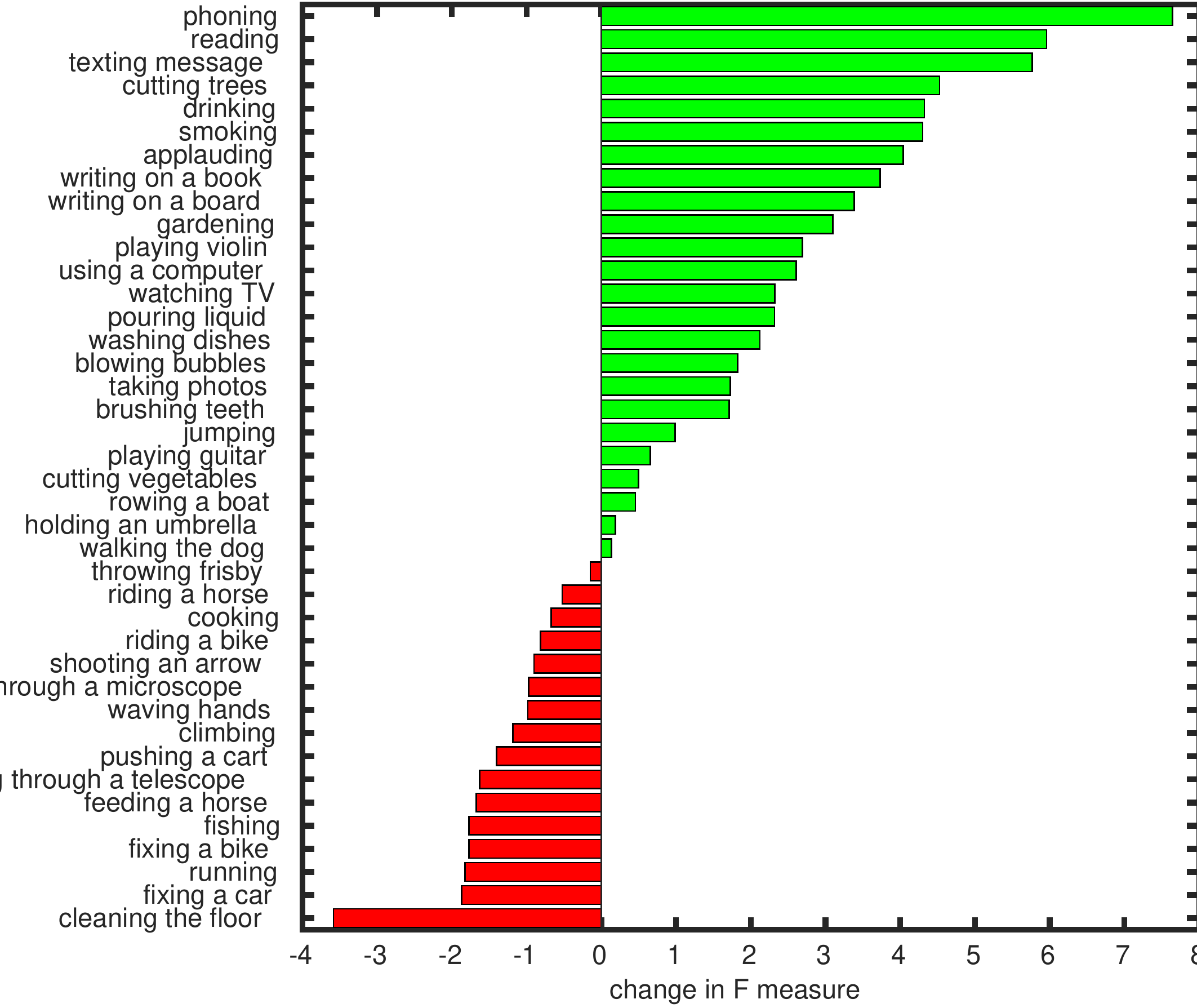}\protect\caption{\label{fig:fine-grained-improvement}Our approach improves mainly
fine-grained tasks and categories where classification depends on
small and specific image windows. The figure shows absolute difference
in terms of F-measure over the baseline approach on all categories
of the Stanford-40 Actions \cite{yao2011human} dataset. It is recommended
to view this figure online. }
\end{figure}

The Standford-40 Action dataset \cite{yao2011human} is a benchmark
dataset made of 9532 images of 40 different action classes, with 4000
images for training and the rest for testing. It contains a diverse
set of action classes including transitive ones with small objects
(smoking, drinking) and large objects (horses), as well as intransitive
actions (running, jumping). As a baseline, we used the GAP-network
of \cite{zhou2015learning} as a feature extractor and trained a multi-class
SVM \cite{liblinear} using the resulting features. It is particularly
interesting to examine the classes for which our method is most beneficial.
We have calculated the F-measure for each class using the classification
scores from the fourth and first iteration and compared them. Fig.
\ref{fig:fine-grained-improvement} shows this; the largest absolute
improvements are on relatively challenging classes such as texting
a message (7.64\%), drinking (4.32\%), smoking (4.3\%), etc. For all
of these, the discriminative objects are small objects and are relatively
hard to detect compared to most other classes. In some cases, performance
is harmed by zooming in on too-local parts of an image: for ``riding
a bike'' (-0.8\%), a small part of the bicycle will not allow disambiguating
the image from e.g., ``fixing a bike''. Another pair of categories
exhibiting similar behavior is ``riding a horse'' vs. ``feeding
a horse''.

\subsection{Top-Down vs. Bottom-Up attention}

To further verify that our introspection mechanism highlights regions
whose exploration is worthwhile, we evaluated an alternative to the
introspection stage by using a generic saliency measure \cite{zhu2014saliency}.
On the Stanford-40 dataset, instead of using the CAM after the first
classification step, we picked the most salient image point as the
center of the next sub-window. Then we proceeded with training and
testing as usual. This produced a sharp drop in results: on the first
iteration performance dropped from 74.47\% when using the CAM to 62.31\%
when using the saliency map. Corresponding drops in performance were
measured in the late-fusion and early fusion steps, which improve
results in the proposed scheme but made them worse when using saliency
as a guide.

\subsubsection{Usage of Complementary Feature Representations}

The network used for drawing attention to discriminative image regions
need not necessarily be the one used for feature representation. We
used the VGG-16 \cite{simonyan2014very} network to extract fc6 features
along the GAP features for all considered windows. On the Stanford-40
Actions dataset, when used to classify categories using features extracted
from entire images, these features we slightly weaker than the GAP
features (73\% vs 75\%). However, training on a concatenated feature
representation boosted results significantly, reaching a precision
of 80\%.\textbf{ }We observed a similar effect on all datasets, showing
that the two representations are complementary in nature. Combined
with our iterative method, we were able to achieve 81.74\%, compared
to the previous best 81\% of \cite{gao2015deep}.

\subsubsection{Effect of Aspect-Ratio Distortion}

Interestingly, our baseline implementation (using only the VGG-GAP
network as a feature extractor for the entire image) got a precision
score of 75.23\% compared to 72.03\% of \cite{zhou2015learning}.
We suspect that it may be because in their implementation , they modified
the aspect ratio of the images to be square regardless of the original
aspect ratio, whereas we did not. Doing so indeed got a score more
similar to theirs, which is interesting from a practical viewpoint.

\subsection{\label{sub:Various-Parameters-=000026}Various Parameters \& Fine-Tuning}

Our method includes several parameters, including the number of iterations,
the width of the beam-search used to explore routes of windows on
the image and the ratio between the size of the current window and
the next. For the number iterations, we have consistently observed
that performance saturates, and even deteriorates a bit, around iteration
4. An example of this can be seen in Fig. \ref{fig:Early-vs.-Late}
showing the performance vs iteration number on the Standford-40 dataset.
A similar behavior was observed on all the datasets on which we've
evaluated the method. This is probably due to the increasingly small
image regions considered at each iteration. As for the number of windows
to consider at each stage, we tried choosing between 1 and 3 of the
windows relating to the highest ranking classes on a validation set.
At best, such strategies performed as well as the greedy strategy,
which chose only the highest scoring window at each iteration. The
size of the sub-window with respect to the current image window was
set as $\sqrt{2}m$ where $m$ is the geometric mean of the current
window's height and width (in effect, all windows are square, except
the entire image). We have experimented with smaller and larger values
on a validation set and found this parameter to give a good trade-off
between not zooming in too much (risking ``missing'' relevant features)
and too little (gaining too little information with respect to the
previous iteration). 

We have also evaluated our results when fine-tuning the VGG-GAP network
before the first iteration. This improves the results for some of
the datasets, i.e., Stanford-40 \cite{yao2011human}, CUB-200-2011
\cite{wah2011caltech}, but did not improve results significantly
for others (Dogs \cite{KhoslaYaoJayadevaprakashFeiFei_FGVC2011},
Aircraft \cite{maji2013fine}). 

Finally, we evaluated the effect of fine-tuning the network for \emph{all
}iterations on the CUB-200-2011. This resulted in a competitive results
of 79.95\%. Some of the best results to date added a massive amount
of external data mined from the web \cite{krause2015unreasonable}
(91.9\%) and/or strong supervision \cite{xu2015augmenting} (84.6\%).

\section{Conclusions\label{sec:Conclusions}}

We have presented a method, which by repeatedly examining the source
of the current prediction, decides on informative image regions to
consider for further examination. The method is based on the observation
that a trained CNN can be used to highlight relevant image areas even
when its final classification is incorrect. This is a result of training
on multiple visual categories using a shared feature representation.
We have built upon Class Activation Maps \cite{zhou2015learning}
due to their simplicity and elegance, though other methods for identifying
the source of the classification decision (e.g., \cite{zeiler2014visualizing})
could probably be employed as well. The proposed method integrates
multiple features extracted at different locations and scales. It
makes consistent improvement over baselines on fine-grained classification
tasks and on tasks where classification depends on fine localized
details. It obtains competitive results on CUB-200-2011 \cite{wah2011caltech},
among methods which avoid strong supervision such as bounding boxes
or keypoint annotations. The improvements are shown despite the method
being trained using only class labels, avoiding the need for supervision
in the form of part annotations or even bounding boxes. In future
work, it would be interesting to examine the use of recurrent nets
(RNN, LSTM \cite{HochreiterSchmidhuber97}) to automatically learn
sequential processes, which incrementally improve classification results,
extending the approach described in the current work. 

\clearpage{}

\bibliographystyle{splncs}
\bibliography{egbib}

\end{document}